\documentclass[twoside,11pt]{article}

\usepackage{jmlr2e}

\usepackage[numbers]{natbib}

\usepackage{amsmath,amssymb, bm}
\usepackage{multirow}
\usepackage{caption}
\newcommand{\tabularspace}{\vspace{3pt}}
\usepackage{cleveref}
\newcommand{\Ltwo}{L^2}
\newcommand{\fp}[1]{\bm{f}_{\bm{#1}}}
\newcommand{\RF}{\mathcal{R}}
\newcommand{\knn}{\mathrm{knn}}
\newcommand{\methodname}{{3D-ST}}

\ShortHeadings{}{}
\firstpageno{1}

\begin{document}

\title{Anomaly Detection in 3D Point Clouds \\ using Deep Geometric Descriptors}

\author{\name Paul Bergmann \email paul.bergmann@mvtec.com \\
       \addr MVTec Software GmbH\\
       Technical University of Munich       \AND
       \name David Sattlegger \email sattlegger@mvtec.com \\
       \addr MVTec Software GmbH}

\editor{}

\maketitle

\begin{abstract}%
We present a new method for the unsupervised detection of geometric anomalies in high-resolution 3D point clouds. In particular, we propose an adaptation of the established student-teacher anomaly detection framework to three dimensions. A student network is trained to match the output of a pretrained teacher network on anomaly-free point clouds. When applied to test data, regression errors between the teacher and the student allow reliable localization of anomalous structures.
To construct an expressive teacher network that extracts dense local geometric descriptors, we introduce a novel self-supervised pretraining strategy. The teacher is trained by reconstructing local receptive fields and does not require annotations. Extensive experiments on the comprehensive MVTec 3D Anomaly Detection dataset highlight the effectiveness of our approach, which outperforms the next-best method by a large margin. Ablation studies show that our approach meets the requirements of practical applications regarding performance, runtime, and memory consumption.
\end{abstract}

\section{Introduction}

In recent years, significant progress has been made in the field of 3D computer vision in various research areas such as 3D classification, 3D semantic segmentation, and 3D object recognition. Many new methods build on earlier achievements in their counterparts in 2D, which operate with natural image data. However, the transition from 2D to 3D poses additional challenges, e.g., the need to deal with unordered point clouds and sensor noise. This has led to the development of new network architectures and training protocols specific to three dimensions.

We consider the challenging task of unsupervised anomaly detection and localization in 3D point clouds. The goal is to detect data points that deviate significantly from a training set of exclusively anomaly-free samples. This problem has important applications in various fields, such as industrial inspection \citep{Bergmann_2021_IJCV,Bergmann_2022_mvtec_3dad,handcrafted_feature_dictionary_nanofibres,song_steel_surface_defect_database}, autonomous driving \citep{fishyscapes2019,hendrycks_benchmark_anomaly_segmentation}, and medical imaging \citep{Bakas_2017,c_baur_vae_gan,brats2015}. It has received considerable attention in 2D, where models are typically trained on color or grayscale images with established and well-studied architectures based on convolutional neural networks. In 3D, this problem is still comparatively unexplored and only a small number of methods exists. In this work, following approaches in other computer vision areas, we draw inspiration from recent advances in 2D anomaly detection to devise a powerful 3D method.

More specifically, we build on the success of using descriptors from pretrained neural networks for unsupervised anomaly detection in 2D. An established protocol is to extract these descriptors as intermediate features from networks trained on the ImageNet \citep{krizhevsky2012imagenet} dataset. Models based on pretrained features were shown to perform better than ones trained with random weight initializations  \citep{bergmann2020uninformed,burlina2019whereiswally,cohen_2020_spade}. In particular, they outperform methods based on convolutional autoencoders or generative adversarial networks. 

\begin{figure}[t]
    \centering
\includegraphics[width=0.95\textwidth]{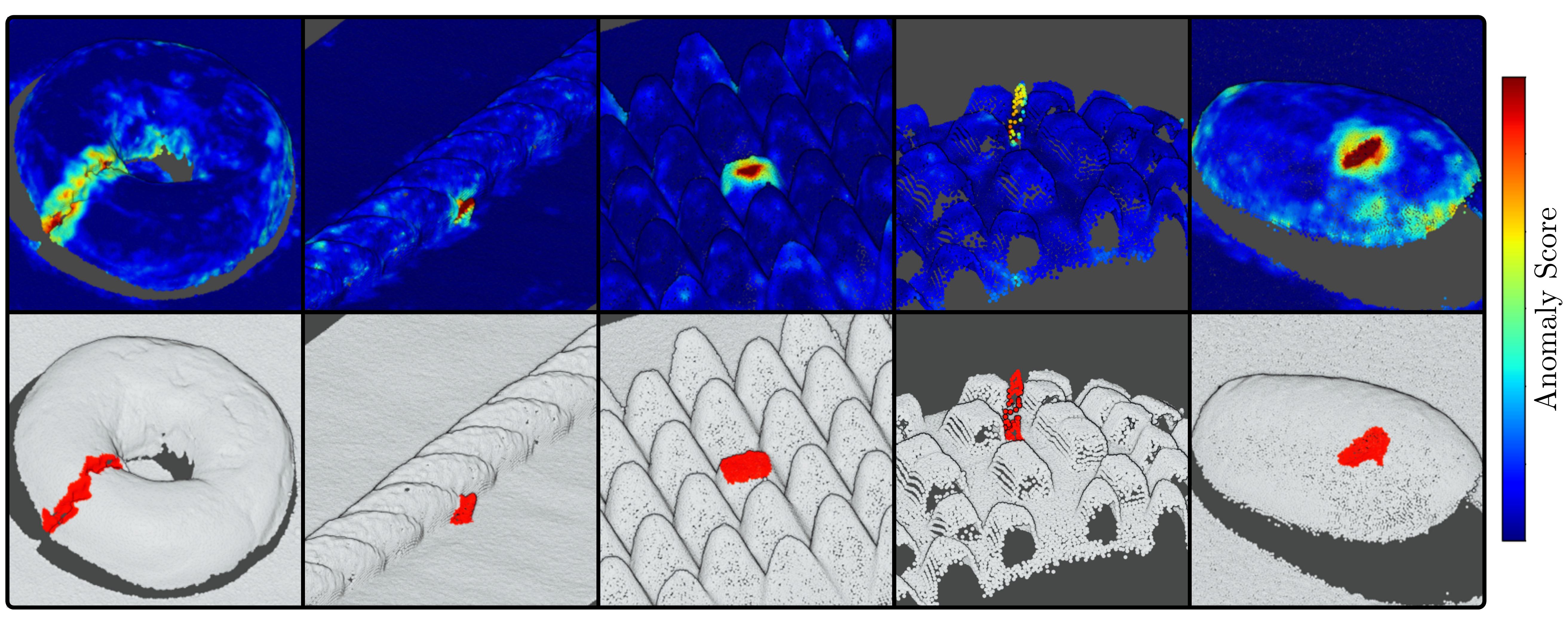}
    \caption{Qualitative results of our method on the MVTec 3D Anomaly Detection dataset. It reliably localizes geometric anomalies in test point clouds, while being only trained on anomaly-free samples. Top row: Anomaly scores for each 3D point predicted by our algorithm. Bottom row: Ground truth annotations of anomalous points in red.}
    \label{fig:teaser_figure}
\end{figure}

So far, there is no established pretraining protocol for unsupervised anomaly detection in 3D point clouds. Existing work addresses the extraction of local 3D features that are highly task-specific. For point cloud registration, feature extractors often heavily downsample the input data or operate only on a small number of input points. This makes them ill-suited for anomaly localization in 3D. 
In this work, we develop a novel approach for pretraining local geometric descriptors that transfer well to this task. We then use this pretraining strategy to introduce a new method that outperforms existing approaches in the localization of geometric anomalies in high-resolution 3D point clouds. In particular, our key contributions are:

\begin{itemize}
    \setlength\itemsep{1ex}
    \item[$\bullet$] We present 3D Student-Teacher (\methodname{}), the first method for unsupervised anomaly detection that operates directly on 3D point clouds. Our method is trained only on anomaly-free data and it localizes geometric anomalies in high-resolution test samples with a single forward pass. We propose an adaptation of a student-teacher framework for anomaly detection to three dimensions. A student network is trained to match deep local geometric descriptors of a pretrained teacher network. During inference, anomaly scores are derived from the regression errors between the student's predictions and the teacher's targets. Our method sets a new state of the art on the recently introduced MVTec 3D-AD dataset. It performs significantly better than existing methods that use voxel grids and depth images.
    
    \item[$\bullet$] We develop a self-supervised training protocol that allows the teacher network to learn generic local geometric descriptors that transfer well to the 3D anomaly detection task. The teacher extracts a geometric descriptor for each input point by aggregating local features within a limited receptive field. A decoder network is trained to reconstruct the local geometry encoded by the descriptors. Our pretraining strategy provides explicit control over the receptive field and dense feature extraction for a large number of input points. This allows us to compute anomaly scores for high-resolution point clouds without the need for intermediate subsampling.
\end{itemize}

\section{Related Work}

Our work touches on several aspects of computer vision, namely unsupervised detection of anomalies in two and three dimensions and extraction of deep local geometric descriptors for 3D data. 

\subsection{Anomaly Detection in 2D}

There is a large body of work on the unsupervised detection of anomalies in two dimensions, i.e., in RGB or grayscale images. \citet{ehret_review_paper_2019} and \citet{pang2021adreview} give comprehensive overviews. Some of the existing methods are trained from scratch with random weight initialization, in particular, those based on convolutional autoencoders (AEs)
\citep{Bergmann_2019_SSIM_AE,Hong_2020_DecentralizationLoss,Liu_2020_VisuallyExplaining,Venkataramanan_2020_AttentionGuided,Wang_2020_LatentSpaceResampling} or generative adversarial networks (GANs) \citep{Carrara_2021_CBiGAN,Potter_2020_pandanet,Schlegl_2019_fAnoGan}. 

A different class of methods leverage descriptors from pretrained networks for anomaly detection \citep{bergmann2020uninformed,cohen_2020_spade,Defard_2021_PaDiM,Gudovskiy_2022_CFLOW,Mishra_2020_piade,Reiss_2021_PANDA,Rippel_2021_Gaussian}. The key idea behind these approaches is that anomalous regions produce descriptors that differ from the ones without anomalies. These methods tend to perform better than methods trained from scratch, which motivates us to transfer this idea to the 3D domain. 

\citet{bergmann2020uninformed} propose a student-teacher framework for 2D anomaly detection. A teacher network is pretrained on the ImageNet dataset to output descriptors represented by feature maps. Each descriptor captures the content of a local region within the input image. For anomaly detection, an ensemble of student networks is trained on anomaly-free images to reproduce the descriptors of the pretrained teacher. During inference, anomalies are detected when the students produce increased regression errors and predictive variances. Closely following this idea, \citet{Salehi_2021_CVPR} train a single student network to match multiple feature maps of a single teacher.

\subsection{Anomaly Detection in 3D}

To date, there are very few methods that address the task of unsupervised anomaly detection in 3D data. None of them leverages the descriptiveness of feature vectors from pretrained networks.

\citet{Simarro_Viana_2021} propose Voxel f-AnoGAN, which is an extension of the 2D f-AnoGAN model \citep{Schlegl_2019_fAnoGan} to 3D voxel grids. A GAN is trained on anomaly-free data samples. Afterwards, an encoder is trained to predict the latent vectors of anomaly-free voxel grids that, when passed through the generator network, reconstruct the input data. During inference, anomaly scores are derived by a per-voxel comparison of the input to the reconstruction. \citet{Bengs_2021_AE_on_MRI} introduce a method based on convolutional autoencoders that also operates on 3D voxel grids. A variational autoencoder is trained to reconstruct input samples through a low-dimensional bottleneck. Again, anomaly scores are derived by comparing each voxel element of the input to its reconstruction.

Recently, \citet{Bergmann_2022_mvtec_3dad} introduced MVTec 3D-AD, a comprehensive dataset for the evaluation of 3D anomaly detection algorithms. 
So far, this is the only public dataset specifically designed for this task. They show that the existing methods do not perform well on challenging high-resolution point clouds and that there is a need for the development of new methods for this task.

\subsection{Learning Deep 3D Descriptors}

Geometric feature extraction is commonly used in 3D applications such as 3D registration or 3D pose estimation. The community has recently shifted from designing hand-crafted descriptors \citep{salti_2014_shot_descriptor,tombari_2010_descriptors} to learning-based approaches. 

One line of work learns low-dimensional descriptors on local 3D patches cropped from larger input point clouds. In 3DMatch \citep{Zeng_2017_3DMatch} and PPFNet \citep{PPF_Net}, supervised metric learning is used to learn embeddings from annotated 3D correspondences. PPF-FoldNet \citep{PPF_Fold_Net} pursues an unsupervised strategy where an autoencoder is trained on point pair features extracted from the local patches. Similarly,  \citet{wadim2016localrgbddescriptors} introduce an autoencoder that is trained on patches of {RGB-D} images to obtain local features. These methods have the disadvantage that a separate patch  needs to be cropped and processed for each feature. This quickly becomes computationally intractable for a large number of points.

To mitigate this problem, recent 3D feature extractors attempt to densely compute features for high-resolution inputs. \citet{choy2019fcgf} propose FCGF, a fully convolutional approach to local geometric feature extraction for 3D registration. They design a network with sparse convolutions to efficiently processes high-resolution voxel data. Given a large number of precisely annotated local correspondences, their approach is trained using contrastive losses that encourage matching local geometries to be close in feature space. PointContrast \citep{PointContrast2020} learns descriptors for 3D registration in a self-supervised fashion and does not rely on human annotations. Correspondences are automatically derived by augmenting a pair of overlapping views from a single 3D scan. While being computationally efficient, these methods require a prior voxelization that can lead to discretization inaccuracies. Furthermore, all of the discussed methods are designed to produce feature spaces that are ideally invariant to 3D rotations of the input data. In unsupervised anomaly detection, however, anomalies can manifest themselves precisely through locally rotated geometric structures. Such differences should therefore be reflected in the extracted feature vectors. This calls for the development of a different pretraining strategy that is sensitive to local rotations.

\section{Student-Teacher Anomaly Detection in Point Clouds}

In this section, we introduce 3D Student-Teacher (\methodname{}), a versatile framework for the unsupervised detection and localization of geometric anomalies in high-resolution 3D point clouds. We build on the recent success of leveraging local descriptors from pretrained networks for anomaly detection and propose an adaptation of the 2D student-teacher method \citep{bergmann2020uninformed} to 3D data.

\begin{figure}[t]
    \centering
\includegraphics[width=0.99\textwidth]{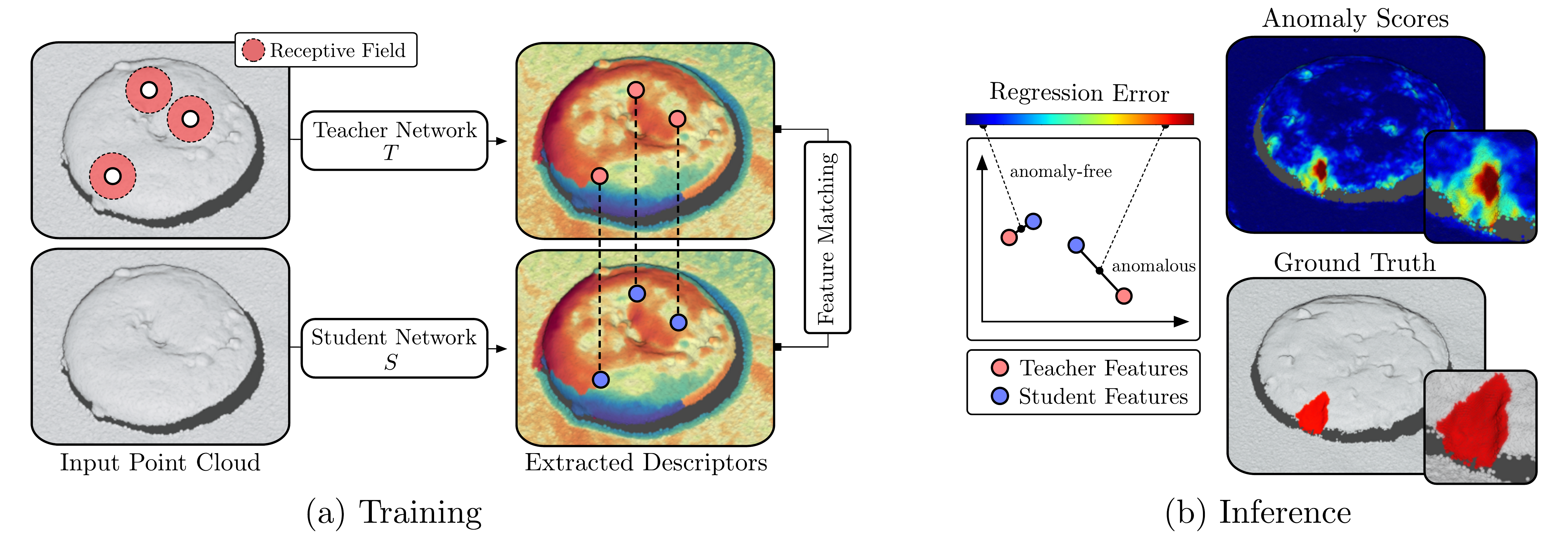}
    \caption{(a) Training of our proposed \methodname{} method on anomaly-free point clouds. A student network $S$ is trained against the local 3D descriptors of a pretrained teacher network $T$. (b) Computation of anomaly scores during inference. Anomaly scores are derived by computing the regression error between the student and the teacher network. Increased regression errors correspond to anomalous 3D points.}
    \label{fig:ad_method_overview}
\end{figure}

Given a training dataset of anomaly-free input point clouds, our goal is to create a model that can localize anomalous regions in test point clouds, i.e., to assign a real-valued anomaly score to each point. To achieve this, we design a dense feature extraction network $T$, called \textit{teacher network}, that computes local geometric features for arbitrary point clouds. For anomaly detection, a \textit{student network} $S$ is trained on the anomaly-free point clouds against the descriptors obtained from $T$. During inference, increased regression errors between $S$ and $T$ indicate anomalous points. An overview of our approach is illustrated in \Cref{fig:ad_method_overview}.

To pretrain the teacher, we present a self-supervised protocol. It works on any generic auxiliary 3D point cloud dataset and requires no human annotations.

\subsection{Self-Supervised Learning of Dense Local Geometric Descriptors}
\label{sec:teacher_pretraining}
\begin{figure}[ht]
    \centering
\includegraphics[width=0.99\textwidth]{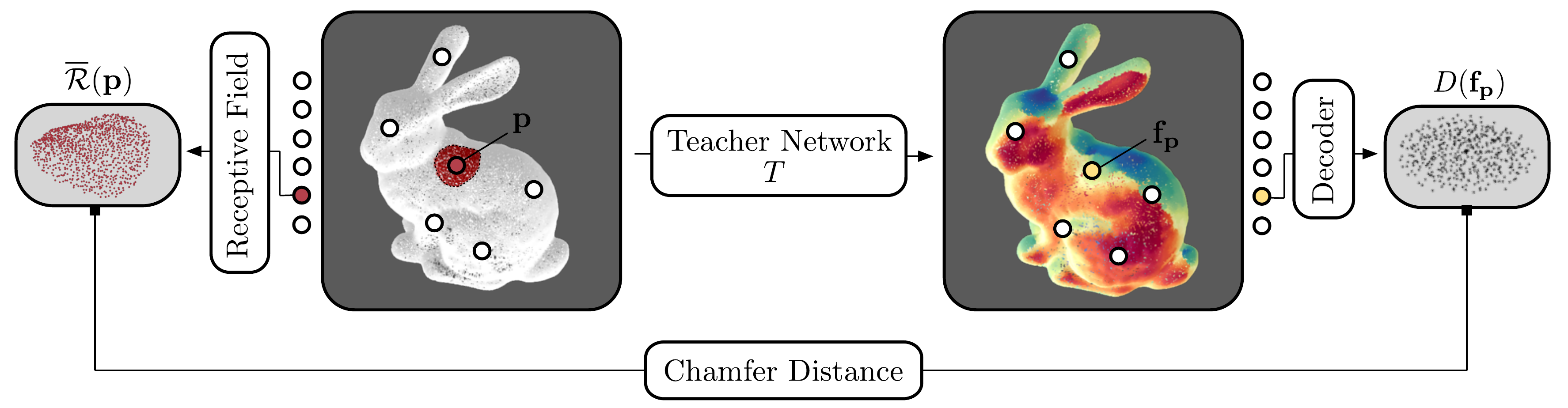}
    \caption{Overview of our proposed self-supervised pretraining strategy. A teacher network is trained to output local geometric descriptors for each 3D point in the input sample with a single forward pass. Simultaneously, a decoder network decodes randomly sampled descriptors of the teacher and attempts to decode the local receptive field around its respective input point.}
    \label{fig:feature_pretraining}
\end{figure}

We begin by describing how to construct a descriptive teacher network $T$. An overview of our pretraining protocol is displayed in \Cref{fig:feature_pretraining}. Given an input point cloud $\bm{P} \subset \mathbb{R}^3$ containing $n$ 3D points, its purpose is to produce a $d$-dimensional feature vector $\fp{p} \in \mathbb{R}^{d}$ for every $\bm{p} \in \bm{P}$.
The vector $\fp{p}$ describes the local geometry around the point $\bm{p}$, i.e., the geometry within its receptive field. 

\paragraph{\textbf{Local Feature Aggregation.}} The network architecture of $T$ has two key requirements. First, it should be able to efficiently process high-resolution point clouds by computing a feature vector for each input point without downsampling the input data. Second, it requires explicit control over the receptive field of the feature vectors. In particular, it has to be possible to efficiently compute all points within the receptive field of an output descriptor. 

To meet these requirements, we construct the $k$-nearest neighbor graph of the input point cloud and initialize $\fp{p} = \bm{0}$. We then pass the input sample through a series of residual blocks, where each block updates the feature vector of each 3D point $\bm{p}$ from $\fp{p} \in \mathbb{R}^d$ to $\tilde{\fp{p}} \in \mathbb{R}^d$. These blocks are inspired by RandLA-Net \citep{hu2019randla,hu2021learning}, an efficient and lightweight neural architecture for semantic segmentation of large-scale point clouds. In semantic segmentation tasks, the absolute position of a point is often related to its class, e.g., in autonomous driving datasets. Here, we want our model to produce features that describe the local geometry of an object independent of its absolute location. We therefore make the residual blocks translation-invariant by removing any dependency on absolute coordinates. This significantly increases the performance when used for anomaly detection as underlined by the results of our experiments in \Cref{sec:experiments}. 

\begin{figure}[t]
    \centering
\includegraphics[width=0.95\textwidth]{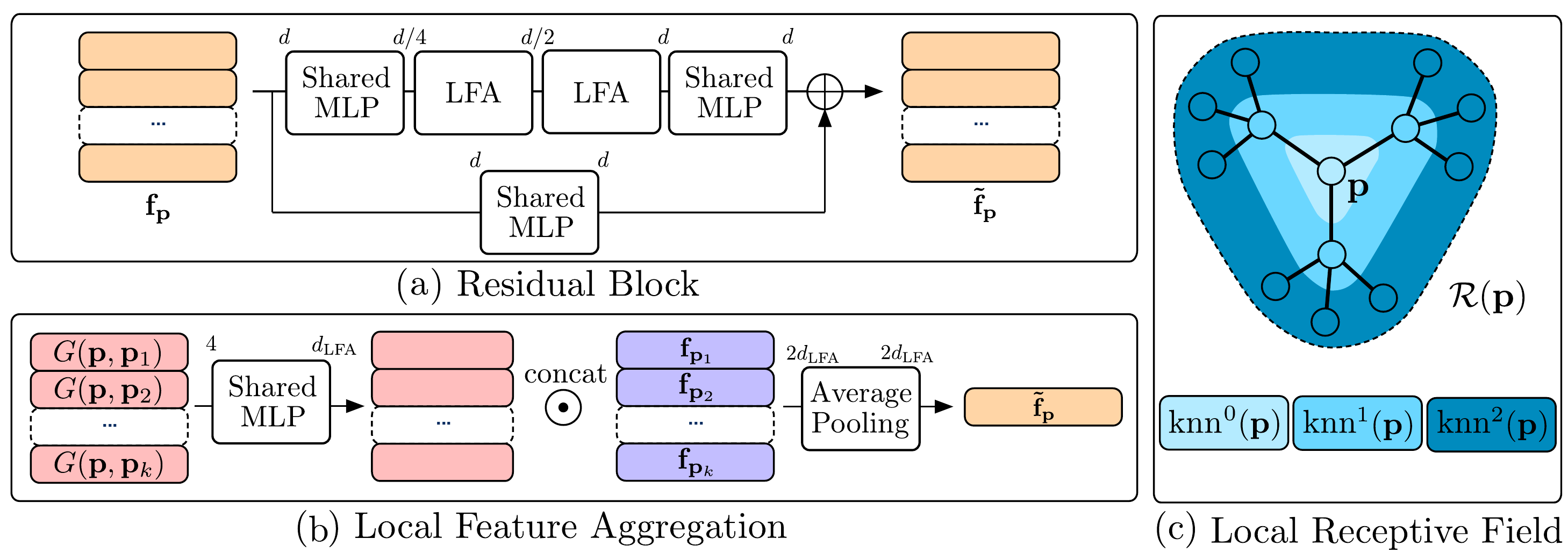}
    \caption{Overview of our network architecture. (a) The local feature aggregation block aggregates geometric information of surrounding points. (b) Residual block that performs a series of local feature aggregation steps to update each points feature vector. (c) Computation of the receptive field for each input point.}
    \label{fig:network_arch}
\end{figure}

The architecture of our residual blocks is visualized in \Cref{fig:network_arch}(a). The input features are first passed through a shared MLP, followed by two local feature aggregation (LFA) blocks. The output features are added to the input after processing both by an additional shared MLP\@. The features are transformed by a series of residual blocks and a final shared MLP with a single hidden layer that maintains the dimension of the descriptors, i.e., $\bm{\tilde f_p} \in \mathbb{R}^d$.

The purpose of the LFA block is to aggregate the geometric information from the local vicinity of each input point. To this end, it computes the nearest neighbors $\knn(\bm{p}) = \{\bm{p}_1, \bm{p}_2, \dots, \bm{p}_k\}$ of all $\bm{p} \in \bm{P}$ and a set of local geometric features $G$ for each point pair defined by
\begin{equation}
G(\bm{p}, \bm{p}_j) = (\bm{p} - \bm{p}_j) \odot \|(\bm{p} - \bm{p}_j)\|_2 ~\textrm{ where } j \in \{1, \dots, k\}.
\end{equation}
The operator $\odot$ denotes the concatenation operation and $\|\cdot\|_2$ denotes the $\Ltwo$-norm. Since $G$ only depends on the difference vectors between neighboring points, our network is by design invariant to translations of the input data. Our experiments show that this invariance of our local feature extractor is crucial for anomaly detection performance. Therefore, we make this small but important change to the LFA block. A schematic description of such a block is given in \Cref{fig:network_arch}(b). 

For each LFA block, the set of geometric features $G(\bm{p}, \bm{p}_j)$ is passed through a shared MLP producing feature vectors of dimension $d_{\mathrm{LFA}}$. These are concatenated with the set of input features $\{ \bm{f}_{\bm{p}_1}, \dots,  \bm{f}_{\bm{p}_k} \}$. The output feature vector of the LFA block $\tilde{\bm{f}}_{\bm{p}}$ is obtained by an average-pooling operation of the concatenated features, yielding a feature vector of dimension $2 d_{\mathrm{LFA}}$. 

\paragraph{\textbf{Reconstructing Local Receptive Fields.}} To pretrain $T$ in a self-supervised fashion, we propose to employ a network $D$ that decodes the local receptive field of a feature vector.

The design of our network architecture allows an efficient computation of all points within the receptive field $\RF(\bm{p})$ of a point $\bm{p}$, i.e., all points that affect the feature vector $\bm{f_p}$. Each LFA block depends on the features of the surrounding nearest neighbors $\knn(\bm{p})$. Whenever an LFA block is executed, $\RF(\bm{p})$ grows by one hop in the nearest-neighbor graph. The receptive field can therefore be obtained by iteratively traversing the nearest neighbor graph:
  \begin{equation}
    \RF(\bm{p}) = \bigcup_{l = 0}^{L} \knn^l(\bm{p})
    , \quad \text{ where } \quad
    \knn^l(\bm{p}) = \bigcup_{\bm{q} \in \knn^{l-1}(\bm{p})} \knn(\bm{q})
  \end{equation}
and $\knn^0 = \{\bm{p}\}$. $L$ denotes the total number of LFA blocks in the network. \Cref{fig:network_arch}(c) visualizes this definition of the receptive field.

The decoder $D: \mathbb{R}^d \rightarrow \mathbb{R}^{3 \times m}$ upsamples a feature vector to produce $m$ 3D points by applying an MLP\@. For pretraining, we extract descriptors from an input point cloud by passing it through the local feature extractor. We then randomly sample a set of points $\bm{Q}$ from the input point cloud. For each $\bm{p} \in \bm{Q}$, we compute the receptive fields $\RF(\bm{p})$ and pass their respective feature vectors through the decoder. To train $D$, we minimize the Chamfer distance \citep{chamfer_distance} between the decoded points and the receptive fields. Since our network architecture is not aware of the absolute coordinates of $\bm{p}$, we additionally compute the mean $\bm{\bar p}$ of all $\bm{p} \in \RF(\bm{p})$ and subtract it from each point, yielding the set $\overline{\RF}(\bm{p}) = \RF(\bm{p}) - \bm{\bar p}$. The loss function for our self-supervised training procedure can then be written as:
\begin{equation}
L_{C}(D) = \frac{1}{|\bm{Q}|} \sum_{\bm{p} \in \bm{Q}} \mathrm{Chamfer}(D(\fp{p}), \overline{\RF}(\bm{p})).
\end{equation}

\paragraph{\textbf{Data Normalization.}}
In order for our teacher network to be applied to any point cloud not included in the pretraining dataset, some form of data normalization is required. Since our network operates on the distance vectors of neighboring points, we choose to normalize the input data with respect to these distances. More specifically, we compute the average distance between each point and its nearest neighbors over the entire training set, i.e.,
\begin{equation}
s = \frac{1}{k \ |\bm{P}|} \ \sum_{\bm{p} \in \bm{P}} \sum_{\bm{q} \in \knn(\bm{p})} \|p - q\|_2.
\label{eqn:scaling_factor}
\end{equation}
We then scale the coordinates of each data sample in the pretraining dataset by $1/s$. This allows us to apply the teacher network to arbitrary point cloud datasets, as long as the same data normalization technique is used. 

\subsection{Matching Geometric Features for 3D Anomaly Detection}

Finally, we describe how to employ the pretrained teacher network $T$ to train a student network $S$ for anomaly detection. Given a dataset of anomaly-free point clouds, we first calculate the scaling factor $s$ for this dataset as defined in \eqref{eqn:scaling_factor}. The weights of $T$ remain constant during the entire anomaly detection training. $S$ exhibits the identical network architecture as $T$ and is initialized with uniformly distributed random weights. Each training point cloud $\bm{P}_t \subset \mathbb{R}^3$ is passed through both networks, $T$ and $S$, to compute dense features $\bm{f}_{\bm{p}}^T$ and $\bm{f}_{\bm{p}}^S$ for all $\bm{p} \in \bm{P}_t$, respectively. The weights of $S$ are optimized to reproduce the geometric descriptors of $T$ by computing the feature-wise $\Ltwo$-distance:
\begin{equation}
    L_{ST}(S) = \frac{1}{|\bm{P}_t|} \sum_{\bm{p} \in \bm{P}_t} \left\|\bm{f}_{\bm{p}}^{S} - (\bm{f}_{\bm{p}}^{T} - \bm{\mu})\ \mathrm{diag}(\bm{\sigma})^{-1} \right\|_2^2.
\end{equation}
We transform the teacher features to be centered around $\bm{0}$ with unit standard deviation. This requires the computation of the component-wise means $\bm{\mu} \in \mathbb{R}^d$ and standard deviations $\bm{\sigma} \in \mathbb{R}^d$ of all teacher features over the whole training set. We denote the inverse of the diagonal matrix filled with the entries of $\bm{\sigma}$ by $\mathrm{diag}(\bm{\sigma})^{-1}$. 

During inference, anomaly scores are derived for each point $\bm{p} \in \bm{P}_i$ in a test point cloud $\bm{P}_i \subset \mathbb{R}^3$. They are given by the regression errors between the respective features of the student and the teacher network, i.e., $\|\bm{f}_{\bm{p}}^S - (\bm{f}_{\bm{p}}^{T} - \bm{\mu})\ \mathrm{diag}(\bm{\sigma})^{-1}\|$. The intuition behind this is that anomalous geometries produce features that the student network has not observed during training, and is hence unable to reproduce. Large regression errors indicate anomalous geometries.

\section{Experiments}
\label{sec:experiments}

To demonstrate the effectiveness of our approach, we perform extensive experiments on the recently released MVTec 3D Anomaly Detection (MVTec 3D-AD) dataset \citep{Bergmann_2022_mvtec_3dad}. This dataset was designed to evaluate methods for the unsupervised detection of geometric anomalies in point cloud data (PCD). Currently, this is the only publicly available comprehensive dataset for this task. It contains over 4000 high-resolution 3D scans of 10 object categories of industrially manufactured products. The task is to train a model on anomaly-free samples and to localize anomalies that occur as defects on the manufactured products during inference.

\subsection{Experiment Setup}

We benchmark the performance of our \methodname{} method against existing methods for unsupervised 3D anomaly detection. In particular, we follow the initial benchmark on MVTec 3D-AD and compare \methodname{} against the Voxel f-AnoGAN, the Voxel Autoencoder, and the Voxel Variation Model. The benchmark also includes their respective counterparts that process depth images instead of voxel grids by exchanging 3D with 2D convolutions. The GAN- and autoencoder-based methods derive anomaly scores by a per-pixel or per-voxel comparison of their reconstructions to the input samples. The Variation Model is a shallow machine learning model that computes the per-pixel or per-voxel means and standard deviations over the training set. During inference, anomaly scores are obtained by computing the per-pixel or per-voxel Mahalanobis distance from a test sample to the training distribution. We employ the same training and evaluation protocols and hyperparameters setting as listed in \citep{Bergmann_2022_mvtec_3dad}.

\paragraph{\textbf{Teacher Pretraining.}}
To pretrain the teacher network of our method (cf.\ \Cref{sec:teacher_pretraining}), we generate synthetic 3D scenes using objects of the ModelNet10 dataset \citep{wu20153d_modelnet10}. It consists of over 5000 3D models divided into 10 different object categories. 

We generate a scene of our pretraining dataset by randomly selecting $10$ samples from ModelNet10 and scaling the longest side of their bounding box to $1$. The objects are rotated around each 3D axis with angles sampled uniformly from the interval $[0, 2 \pi]$. Each object is placed at a random location sampled uniformly from $[-3, 3]^3$. Point clouds are created by selecting $n$ points from the scene using farthest point sampling  \citep{moenning2003_farthest_point_sampling}. The training and validation dataset consist of $1000$ and $50$ point clouds, respectively.
Our experiments show that using such a synthetic dataset for pretraining yields local descriptors that are well suited for 3D anomaly detection. In our ablation studies, we additionally investigate the use of real-world datasets from different domains for pretraining, namely Semantic KITTI \citep{Behley_2019_semantic_kitti,geiger2012cvpr}, MVTec ITODD \citep{Drost2017Itodd}, and 3DMatch \citep{Zeng_2017_3DMatch}.

The teacher network $T$ consists of $4$ residual blocks and processes $n = 64000$ input points. We perform experiments using two different feature dimensions $d \in \{64, 128\}$. The shared MLPs in all network blocks are implemented with a single dense layer, followed by a LeakyReLU activation with a negative slope of $0.2$. The input and output dimensions of each shared MLP are given in \Cref{fig:network_arch}. For local feature aggregation, a nearest neighbor graph with $k=32$ neighbors is constructed. The pretraining runs for $250$ epochs using the Adam optimizer with an initial learning rate of $10^{-3}$ and a weight decay of $10^{-6}$. At each training step, a single input sample is fed through the teacher network. To generate reconstructions of local receptive fields, $16$ randomly selected descriptors from the output of $T$ are passed through the decoder network $D$, which is implemented as an MLP with input dimension $d$, two hidden layers of dimension $128$, and an output layer that reconstructs $m = 1024$ points. Each hidden layer is followed by a LeakyReLU activation with negative slope of $0.05$. After the training, we select the model with the lowest validation error as the teacher network.

\begin{table*}[ht]
    \centering
    \resizebox{0.99\linewidth}{!}{%
    \begin{tabular}{cc|cccccccccc|c} \hline
    & & bagel & \begin{tabular}[c]{@{}c@{}}cable\\ gland\end{tabular} & carrot & cookie & dowel & foam & peach & potato & rope & tire & mean \\ \hline \hline
 \multirow{3}{*}{\rotatebox[origin=c]{90}{Voxel}} & GAN & 0.440 & 0.453  & 0.825 & 0.755 & 0.782 & 0.378 & 0.392 & 0.639 & 0.775 & 0.389 & 0.583 \\
   & AE & 0.260 & 0.341  & 0.581 & 0.351 & 0.502 & 0.234 & 0.351 & 0.658 & 0.015 & 0.185 & 0.348 \\
   & VM & 0.453 & 0.343  & 0.521 & 0.697 & 0.680 & 0.284 & 0.349 & 0.634 & 0.616 & 0.346 & 0.492 \\ \cline{1-13} 
    \multirow{3}{*}{\rotatebox[origin=c]{90}{Depth}} & GAN & 0.111 & 0.072  & 0.212 & 0.174 & 0.160 & 0.128 & 0.003 & 0.042 & 0.446 & 0.075 & 0.143 \\
   & AE & 0.147 & 0.069  & 0.293 & 0.217 & 0.207 & 0.181 & 0.164 & 0.066 & 0.545 & 0.142 & 0.203 \\
   & VM &  0.280 & 0.374  & 0.243 & 0.526 & 0.485 & 0.314 & 0.199 & 0.388 & 0.543 & 0.385 & 0.374 \\ \hline
   \multirow{2}{*}{\rotatebox[origin=c]{90}{PCD}} & \methodname{}$_{64}$ &  0.939 & 0.440  & 0.984 & 0.904 & 0.876 & \textbf{0.633} & 0.937 & \textbf{0.989} & 0.967 & 0.507 & 0.818 \\
   & \methodname{}$_{128}$ &  \textbf{0.950} & \textbf{0.483}  & \textbf{0.986} & \textbf{0.921} & \textbf{0.905} & 0.632 & \textbf{0.945} & 0.988 & \textbf{0.976} & \textbf{0.542} &\textbf{0.833} \\ \cline{1-13}

    \end{tabular}
    }
    \tabularspace
    \caption{Anomaly detection results for each evaluated method and dataset category. The area under the PRO curve is reported for an integration limit of $0.3$. The best performing method is highlighted in boldface.}
    \label{table:localization_results_quantitative}
\end{table*}

\paragraph{\textbf{Anomaly Detection.}}
The student network $S$ in our \methodname{} method has the same network architecture as the teacher. It is trained for $100$ epochs on the anomaly-free training split of the MVTec 3D-AD dataset. We train with a batch size of $1$. This is equivalent to processing a large number of local patches per iteration due to the limited receptive field of the employed networks. We use Adam with an initial learning rate of $10^{-3}$ and weight decay $10^{-5}$. Each point cloud is reduced to $n = 64000$ input points using farthest point sampling. For inference, we select the student network with the lowest validation error.

The evaluation on MVTec 3D-AD requires to predict an anomaly score for each pixel in the original $(x,y,z)$ images. To do this, we apply harmonic interpolation \citep{evans2010harmonicinterp} to the pixels that were not assigned anomaly scores by our method. We follow the standard evaluation protocol of MVTec 3D-AD and compute the per-region overlap (PRO) \citep{Bergmann_2021_IJCV} and the corresponding false positive rate for successively increasing anomaly thresholds. We then report the area under the PRO curve (AU-PRO) integrated up to a false positive rate of $30\%$. We normalize the resulting values to the interval $[0,1]$.

\subsection{Experiment Results}

\Cref{table:localization_results_quantitative} shows quantitative results of each evaluated method on every object category of MVTec 3D-AD. The top three rows list the performance of the voxel-based methods. The following three rows list the performance of the respective methods on 2D depth images. The bottom two rows show the performance of our \methodname{} method on 3D point cloud data, evaluated for two different descriptor dimensions $d \in \{64, 128\}$. Our method performs significantly better than all other methods on every dataset category. Increasing the descriptor dimension from $64$ to $128$ yields a slight overall improvement of $1.5$ percentage points. The latter outperforms the previously leading method by $25.0$ points.

Qualitative results of our method are shown in \Cref{fig:teaser_figure}. \methodname{} manages to localize anomalies over a range of different object categories, such as the crack in the \textit{bagel}, the contamination on the \textit{rope} and the \textit{tire}, or the cut in the \textit{foam} and the \textit{potato}. Additional qualitative results for each object category are shown in \Cref{appendix:qualitatives}.

\begin{figure}[ht]
    \centering
    \includegraphics[width=0.75\textwidth]{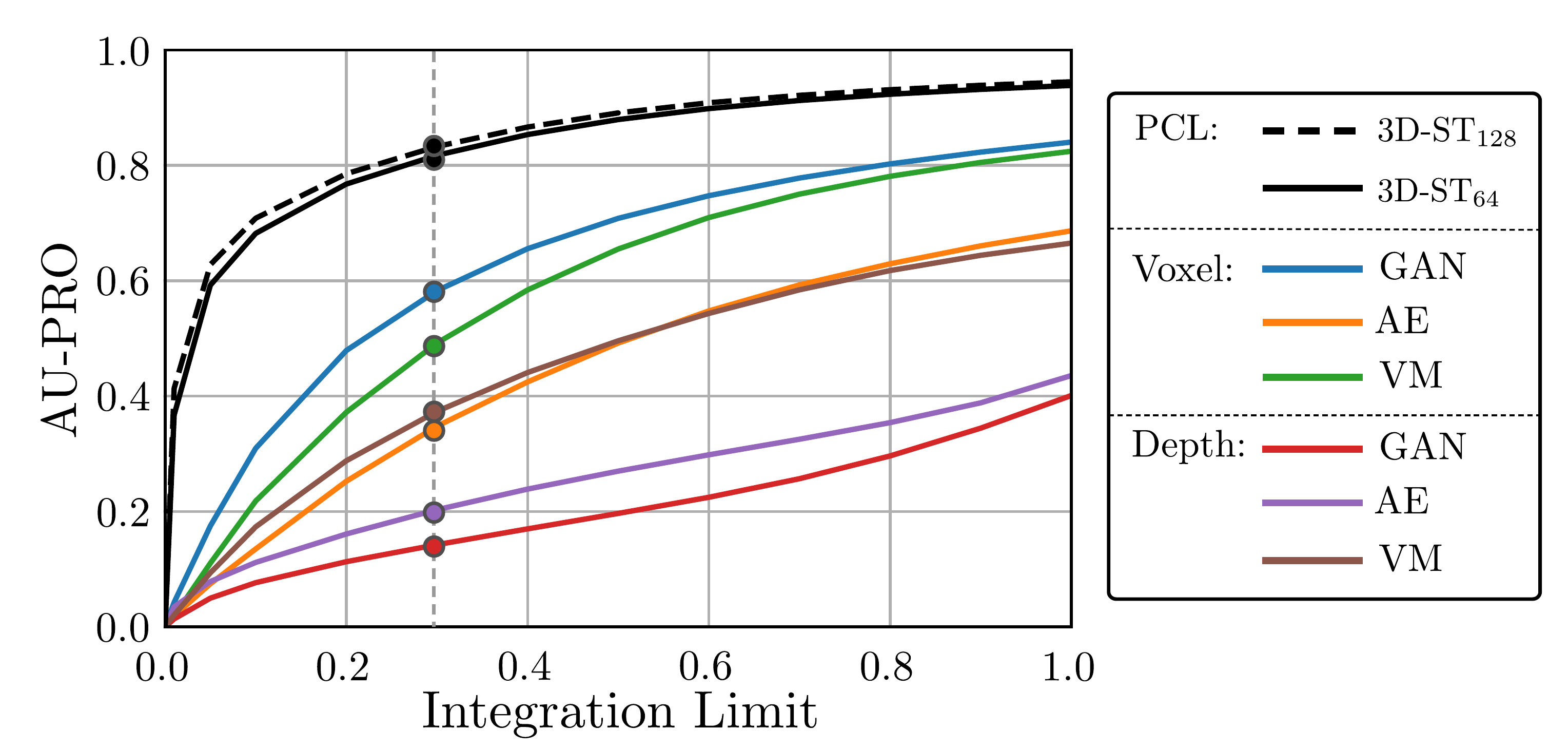}
    \caption{Anomaly detection performance of each evaluated method for varying integration limits. The performance at an integration limit of 0.3 is marked by a vertical line. In real-world scenarios, the performance at lower integration limits is of particular importance.}
    \label{fig:ablation_integration_limit}
\end{figure}

The MVTec 3D-AD paper states that real-world anomaly detection applications require particularly low false positive rates. We therefore report the mean performance of all evaluated methods when varying the integration limit of the PRO curve in \Cref{fig:ablation_integration_limit}. Our method outperforms all other evaluated methods for any chosen integration limit. The relative difference in performance is particularly large for lower integration limits. This makes our approach well-suited for practical applications. Exact values for several integration limits can be found in \Cref{appendix:ablation}.

\subsection{Ablation Studies}

We additionally perform various ablation studies with respect to the key hyperparameters of our proposed method. Again, the exact values for each experiment can be found in \Cref{appendix:ablation}. \Cref{fig:ablation_study} shows the dependency of the mean performance of our method on the number of input points $n$, the feature dimension $d$, or the number of nearest neighbor points $k$ used for local feature aggregation. We additionally visualize the inference time and the memory consumption of each model during training and evaluation\footnote{All models were implemented using the PyTorch library \citep{Paszke_2019_PyTorch}. Inference times and memory consumption were measured on an NVIDIA Tesla V100 GPU.}. We find that our method is insensitive to the choice of each hyperparameter. In particular, the mean performance of each evaluated model outperforms the best performing competing model from the baseline experiments by a large margin. The mean performance of our model grows monotonically with respect to each considered hyperparameter. It eventually saturates, whereas the inference time and memory consumption continue to increase super-linearly.

\begin{figure}[t]
    \centering
    \includegraphics[width=0.95\textwidth]{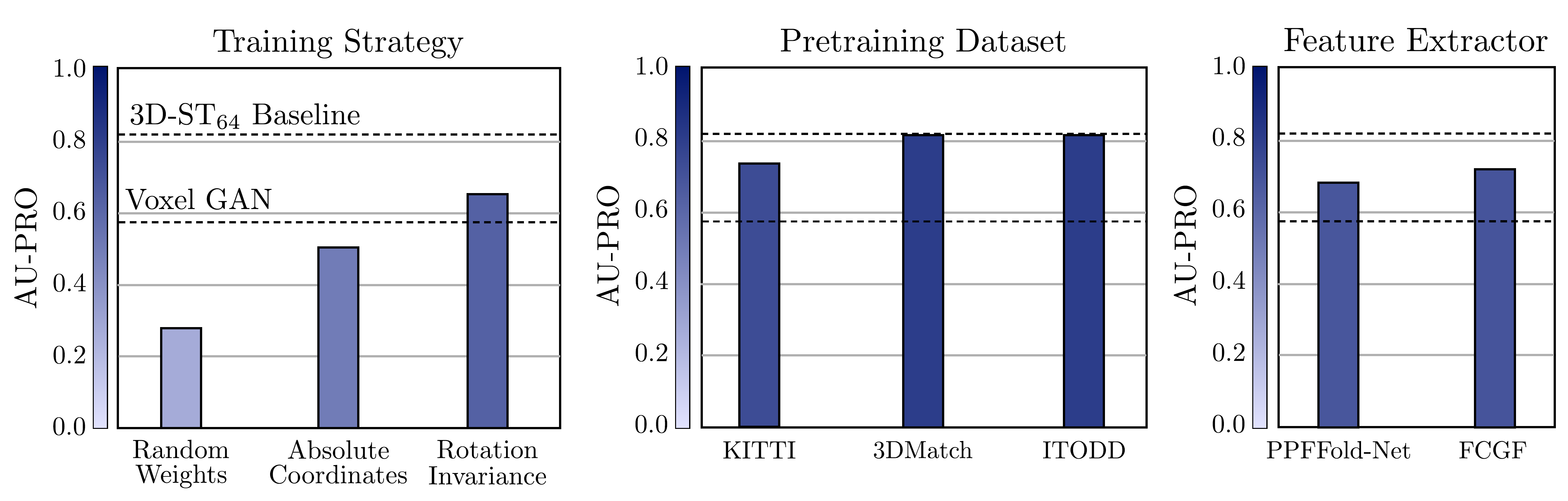}
    \caption{Performance of our method when making changes to the pretraining strategy of the teacher network $T$. We experiment with a random teacher initialization, adding absolute point coordinates, rotation invariant features, and different pretraining datasets.}
    \label{fig:ablation_barplot}
\end{figure}

\paragraph{\textbf{Feature Space of the Teacher Network.}} We depict the effectiveness of our pretraining strategy in \Cref{fig:ablation_barplot}. The left bar plot shows the mean performance with respect to changes in the training strategy of our method. The first bar indicates how the performance changes when we initialize the teacher's weights randomly and perform no pretraining. As expected, the performance drops significantly. The second bar shows the performance when concatenating the absolute point coordinates of each 3D point to the local feature aggregation function $G$, as proposed in \citep{hu2019randla}.
This no longer makes our network translation invariant and decreases the performance. This indicates that translation invariance is indeed important for our network architecture and that our modification to the local feature aggregation module has a significant impact. The third bar shows the performance of our method when trying to additionally incorporate rotation invariance. We achieve this by augmentation of the training data with randomly sampled rotations such that locally rotated geometries are also considered as anomaly-free. In this setting, the performance is still significantly below our method, which indicates that sensitivity to local rotations is beneficial for 3D anomaly detection.

\begin{figure}[ht]
    \centering
\includegraphics[width=0.99\textwidth]{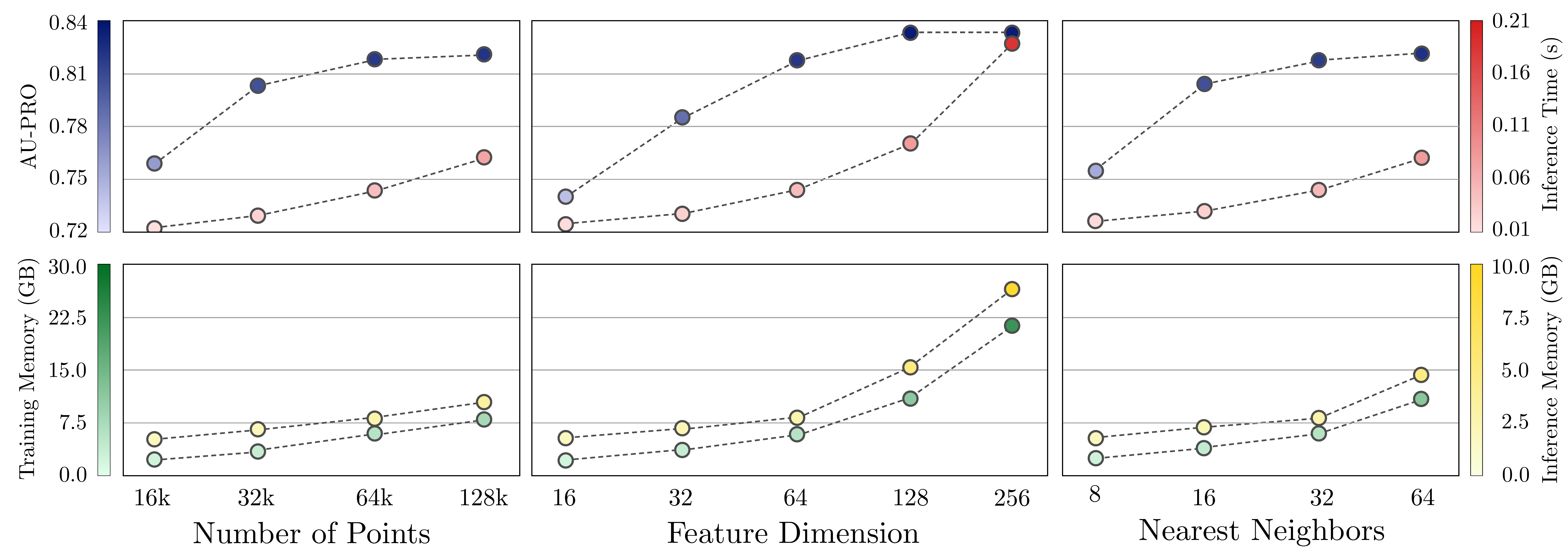}
    \caption{Performance of our method when changing various key hyperparameters, i.e., the number of input points $p$, the feature dimension $d$ of the student and teacher networks, and the number of nearest neighbors $k$ used for local feature aggregation.}
    \label{fig:ablation_study}
\end{figure}

\paragraph{\textbf{Pretraining Dataset.}} In most of our experiments, we use synthetically generated scenes created from objects from the ModelNet10 dataset as described above. Our pretraining strategy does not require any human annotations and can operate on arbitrary input point clouds. We are thus interested in whether the performance varies when using different pretraining datasets. 
As first experiment, we randomly select $1000$ training scenes from the Semantic KITTI autonomous driving dataset, which is captured with a LIDAR sensor. Secondly, we pretrain a teacher on all samples of 3DMatch, an indoor dataset originally designed for point cloud registration. Finally, we use all samples of the MVTec ITODD dataset, an industrial dataset originally designed for 3D pose estimation. 
The center bar chart in  \Cref{fig:ablation_barplot} shows the mean performance of our method when using these three datasets for pretraining, compared to our baseline model. 
We find that our method does not strongly depend on the specific dataset chosen for pretraining when using ITODD or 3DMatch. We observe a slight performance gap for the KITTI dataset, which is likely due to the large domain shift.

\paragraph{\textbf{Feature Extractor.}} 
We additionally test the performance of our method when our pretrained teacher network is replaced by descriptors obtained from different feature extractors.
In particular, we compare against features obtained from PPFFold-Net \citep{PPF_Fold_Net} and FCGF \citep{choy2019fcgf}. For both, we use publicly available pretrained models.\footnote{\url{https://github.com/XuyangBai/PPF-FoldNet}\\\url{https://github.com/chrischoy/FCGF}} 
PPF-FoldNet outputs a single $512$-dimensional descriptor for patches cropped from a local neighborhood of $1024$ points around each input point. Since it requires patch-based feature extraction, producing descriptors for a large number of input points becomes prohibitively slow. We therefore only extract $1000$ descriptors for each point cloud with PPF-FoldNet.
FCGF outputs $32$-dimensionsal descriptors and was pretrained in a supervised fashion on the 3DMatch dataset by finding correspondences for 3D registration. Since it requires a prior voxelization of the input data, we select a voxel size of $3.5\,$mm and extract descriptors for $64000$ points. 

We train our student network to match the features extracted from these pretrained networks instead of our proposed teacher network. The feature dimension of the output layer of our student network is adapted to match the feature dimension of the descriptors. The results are shown in the right bar plot in \Cref{fig:ablation_barplot}. Transferring the features of both networks yields better performance than the Voxel GAN, which is the previously best-performing method that was trained from scratch. This underlines the effectiveness of using pretrained geometric descriptors for 3D anomaly detection. Both extractors do not reach the performance of our proposed pretraining strategy that is specifically designed for the anomaly detection problem.

\section{Conclusion}
We propose \methodname{}, a new method for the challenging problem of unsupervised anomaly detection in 3D point clouds. Our approach is trained exclusively on anomaly-free samples. During inference, it localizes geometric structures that deviate from the ones present in the training set. 
Existing methods such as convolutional autoencoders or generative adversarial networks are trained from random weight initializations.
In contrast to this, our method leverages the descriptiveness of deep local geometric features extracted from a network pretrained on an auxiliary 3D dataset. 
In particular, we propose an adaptation of student-teacher anomaly detection from 2D to 3D. To address the lack of pretraining protocols for 3D anomaly detection, we introduce a self-supervised strategy. This way, we create teacher networks that produce dense local geometric descriptors for arbitrary 3D point clouds. The teacher network is pretrained by reconstructing local receptive fields. For anomaly detection, a student network matches the geometric descriptors of the teacher on anomaly-free data. During inference, anomaly scores are derived for each 3D point by computing the regression error between its associated student and teacher descriptors. Extensive experiments on the MVTec 3D Anomaly Detection dataset show that our method outperforms all existing methods by a large margin. We performed various ablation studies that additionally showed that our method is computationally efficient, and robust to the choice of hyperparameters and pretraining datasets used.

\paragraph{Acknowledgements.} We would like to thank Bertram Drost, Carsten Steger, Markus Glitzner, and the entire research team at MVTec Software GmbH for valuable discussions.

\vskip 0.2in
\bibliographystyle{abbrvnat}
\bibliography{sample}

\begin{thebibliography}{49}
\providecommand{\natexlab}[1]{#1}
\providecommand{\url}[1]{\texttt{#1}}
\expandafter\ifx\csname urlstyle\endcsname\relax
  \providecommand{\doi}[1]{doi: #1}\else
  \providecommand{\doi}{doi: \begingroup \urlstyle{rm}\Url}\fi

\bibitem[Bakas et~al.(2017)Bakas, Akbari, Sotiras, Bilello, Rozycki, Kirby,
  et~al.]{Bakas_2017}
S.~Bakas, H.~Akbari, A.~Sotiras, M.~Bilello, M.~Rozycki, J.~S. Kirby, et~al.
\newblock Advancing the cancer genome atlas glioma {MRI} collections with
  expert segmentation labels and radiomic features.
\newblock \emph{Scientific Data}, 4\penalty0 (1), 2017.
\newblock \doi{10.1038/sdata.2017.117}.

\bibitem[Barrow et~al.(1977)Barrow, Tenenbaum, Bolles, and
  Wolf]{chamfer_distance}
H.~G. Barrow, J.~M. Tenenbaum, R.~C. Bolles, and H.~C. Wolf.
\newblock Parametric correspondence and chamfer matching: Two new techniques
  for image matching.
\newblock In \emph{IJCAI}, pages 659--663, 1977.

\bibitem[Baur et~al.(2019)Baur, Wiestler, Albarqouni, and
  Navab]{c_baur_vae_gan}
C.~Baur, B.~Wiestler, S.~Albarqouni, and N.~Navab.
\newblock {Deep Autoencoding Models for Unsupervised Anomaly Segmentation in
  Brain MR Images}.
\newblock In A.~Crimi, S.~Bakas, H.~Kuijf, F.~Keyvan, M.~Reyes, and T.~van
  Walsum, editors, \emph{Brainlesion: Glioma, Multiple Sclerosis, Stroke and
  Traumatic Brain Injuries}, pages 161--169, Cham, 2019. Springer International
  Publishing.

\bibitem[Behley et~al.(2019)Behley, Garbade, Milioto, Quenzel, Behnke,
  Stachniss, and Gall]{Behley_2019_semantic_kitti}
J.~Behley, M.~Garbade, A.~Milioto, J.~Quenzel, S.~Behnke, C.~Stachniss, and
  J.~Gall.
\newblock {SemanticKITTI: A Dataset for Semantic Scene Understanding of LiDAR
  Sequences}.
\newblock In \emph{Proc. of the IEEE/CVF International Conf.~on Computer Vision
  (ICCV)}, 2019.

\bibitem[Bengs et~al.(2021)Bengs, Behrendt, Krüger, Opfer, and
  Schlaefer]{Bengs_2021_AE_on_MRI}
M.~Bengs, F.~Behrendt, J.~Krüger, R.~Opfer, and A.~Schlaefer.
\newblock {Three-dimensional deep learning with spatial erasing for
  unsupervised anomaly segmentation in brain {MRI}}.
\newblock \emph{International Journal of Computer Assisted Radiology and
  Surgery}, 16, 2021.
\newblock \doi{10.1007/s11548-021-02451-9}.

\bibitem[Bergmann et~al.(2019)Bergmann, L{\"o}we, Fauser, Sattlegger, and
  Steger]{Bergmann_2019_SSIM_AE}
P.~Bergmann, S.~L{\"o}we, M.~Fauser, D.~Sattlegger, and C.~Steger.
\newblock {Improving Unsupervised Defect Segmentation by Applying Structural
  Similarity to Autoencoders}.
\newblock In A.~Tremeau, G.~Farinella, and J.~Braz, editors, \emph{14th
  International Joint Conference on Computer Vision, Imaging and Computer
  Graphics Theory and Applications}, volume 5: VISAPP, pages 372--380,
  Set{\'u}bal, 2019. Scitepress.

\bibitem[Bergmann et~al.(2020)Bergmann, Fauser, Sattlegger, and
  Steger]{bergmann2020uninformed}
P.~Bergmann, M.~Fauser, D.~Sattlegger, and C.~Steger.
\newblock Uninformed students: Student-teacher anomaly detection with
  discriminative latent embeddings.
\newblock In \emph{2020 IEEE/CVF Conference on Computer Vision and Pattern
  Recognition (CVPR)}, pages 4182--4191, 2020.

\bibitem[Bergmann et~al.(2021)Bergmann, Batzner, Fauser, Sattlegger, and
  Steger]{Bergmann_2021_IJCV}
P.~Bergmann, K.~Batzner, M.~Fauser, D.~Sattlegger, and C.~Steger.
\newblock {The MVTec Anomaly Detection Dataset: A Comprehensive Real-World
  Dataset for Unsupervised Anomaly Detection}.
\newblock \emph{International Journal of Computer Vision}, 129\penalty0
  (4):\penalty0 1038--1059, 2021.
\newblock \doi{10.1007/s11263-020-01400-4}.

\bibitem[Bergmann et~al.(2022)Bergmann, Jin, Sattlegger, and
  Steger]{Bergmann_2022_mvtec_3dad}
P.~Bergmann, X.~Jin, D.~Sattlegger, and C.~Steger.
\newblock {The MVTec 3D-AD Dataset for Unsupervised 3D Anomaly Detection and
  Localization}.
\newblock In \emph{17th International Joint Conference on Computer Vision,
  Imaging and Computer Graphics Theory and Applications}, volume 5: VISAPP,
  Set{\'u}bal, 2022. Scitepress.

\bibitem[Blum et~al.(2019)Blum, Sarlin, Nieto, Siegwart, and
  Cadena]{fishyscapes2019}
H.~Blum, P.-E. Sarlin, J.~Nieto, R.~Siegwart, and C.~Cadena.
\newblock {Fishyscapes: A Benchmark for Safe Semantic Segmentation in
  Autonomous Driving}.
\newblock In \emph{2019 IEEE/CVF International Conference on Computer Vision
  Workshop (ICCVW)}, pages 2403--2412, 2019.
\newblock \doi{10.1109/ICCVW.2019.00294}.

\bibitem[Burlina et~al.(2019)Burlina, Joshi, and Wang]{burlina2019whereiswally}
P.~Burlina, N.~Joshi, and I.-J. Wang.
\newblock {Where's Wally Now? Deep Generative and Discriminative Embeddings for
  Novelty Detection}.
\newblock In \emph{IEEE Conference on Computer Vision and Pattern Recognition
  (CVPR)}, 2019.

\bibitem[Carrara et~al.(2021)Carrara, Amato, Brombin, Falchi, and
  Gennaro]{Carrara_2021_CBiGAN}
F.~Carrara, G.~Amato, L.~Brombin, F.~Falchi, and C.~Gennaro.
\newblock Combining {GANs} and {AutoEncoders} for efficient anomaly detection.
\newblock In \emph{2020 25th International Conference on Pattern Recognition
  ({ICPR})}. {IEEE}, 2021.
\newblock \doi{10.1109/icpr48806.2021.9412253}.

\bibitem[Carrera et~al.(2017)Carrera, Manganini, Boracchi, and
  Lanzarone]{handcrafted_feature_dictionary_nanofibres}
D.~Carrera, F.~Manganini, G.~Boracchi, and E.~Lanzarone.
\newblock {Defect Detection in {SEM} Images of Nanofibrous Materials}.
\newblock \emph{IEEE Transactions on Industrial Informatics}, 13\penalty0
  (2):\penalty0 551--561, 2017.
\newblock \doi{10.1109/TII.2016.2641472}.

\bibitem[Choy et~al.(2019)Choy, Park, and Koltun]{choy2019fcgf}
C.~Choy, J.~Park, and V.~Koltun.
\newblock Fully convolutional geometric features.
\newblock In \emph{2019 IEEE/CVF International Conference on Computer Vision
  (ICCV)}, pages 8957--8965, 2019.
\newblock \doi{10.1109/ICCV.2019.00905}.

\bibitem[Cohen and Hoshen(2020)]{cohen_2020_spade}
N.~Cohen and Y.~Hoshen.
\newblock Sub-image anomaly detection with deep pyramid correspondences.
\newblock arXiv preprint arXiv:2005.02357, 2020.

\bibitem[Defard et~al.(2021)Defard, Setkov, Loesch, and
  Audigier]{Defard_2021_PaDiM}
T.~Defard, A.~Setkov, A.~Loesch, and R.~Audigier.
\newblock Padim: A patch distribution modeling framework for anomaly detection
  and localization.
\newblock In A.~Del~Bimbo, R.~Cucchiara, S.~Sclaroff, G.~M. Farinella, T.~Mei,
  M.~Bertini, H.~J. Escalante, and R.~Vezzani, editors, \emph{Pattern
  Recognition. ICPR International Workshops and Challenges}, pages 475--489.
  Springer International Publishing, 2021.
\newblock ISBN 978-3-030-68799-1.

\bibitem[Deng et~al.(2018{\natexlab{a}})Deng, Birdal, and Ilic]{PPF_Fold_Net}
H.~Deng, T.~Birdal, and S.~Ilic.
\newblock {PPF-FoldNet: Unsupervised Learning of Rotation Invariant 3D Local
  Descriptors}.
\newblock In \emph{Proceedings of the European Conference on Computer Vision
  (ECCV)}, 2018{\natexlab{a}}.

\bibitem[Deng et~al.(2018{\natexlab{b}})Deng, Birdal, and Ilic]{PPF_Net}
H.~Deng, T.~Birdal, and S.~Ilic.
\newblock {PPFNet: Global Context Aware Local Features for Robust 3D Point
  Matching}.
\newblock In \emph{Proceedings of the IEEE Conference on Computer Vision and
  Pattern Recognition (CVPR)}, 2018{\natexlab{b}}.

\bibitem[Drost et~al.(2017)Drost, Ulrich, Bergmann, H\"artinger, and
  Steger]{Drost2017Itodd}
B.~Drost, M.~Ulrich, P.~Bergmann, P.~H\"artinger, and C.~Steger.
\newblock {Introducing MVTec ITODD — A Dataset for 3D Object Recognition in
  Industry}.
\newblock In \emph{IEEE International Conference on Computer Vision Workshops
  (ICCVW)}, pages 2200--2208, 2017.
\newblock \doi{10.1109/ICCVW.2017.257}.

\bibitem[Ehret et~al.(2019)Ehret, Davy, Morel, and
  Delbracio]{ehret_review_paper_2019}
T.~Ehret, A.~Davy, J.-M. Morel, and M.~Delbracio.
\newblock {Image Anomalies: A Review and Synthesis of Detection Methods}.
\newblock \emph{Journal of Mathematical Imaging and Vision}, 61\penalty0
  (5):\penalty0 710--743, 2019.

\bibitem[Evans(2010)]{evans2010harmonicinterp}
L.~C. Evans.
\newblock \emph{Partial differential equations}.
\newblock American Mathematical Society, Providence, R.I., 2010.
\newblock ISBN 9780821849743 0821849743.

\bibitem[Geiger et~al.(2012)Geiger, Lenz, and Urtasun]{geiger2012cvpr}
A.~Geiger, P.~Lenz, and R.~Urtasun.
\newblock {Are we ready for Autonomous Driving? The KITTI Vision Benchmark
  Suite}.
\newblock In \emph{Proc.~of the IEEE Conf.~on Computer Vision and Pattern
  Recognition (CVPR)}, pages 3354--3361, 2012.

\bibitem[Gudovskiy et~al.(2022)Gudovskiy, Ishizaka, and
  Kozuka]{Gudovskiy_2022_CFLOW}
D.~Gudovskiy, S.~Ishizaka, and K.~Kozuka.
\newblock {CFLOW-AD: Real-Time Unsupervised Anomaly Detection With Localization
  via Conditional Normalizing Flows}.
\newblock In \emph{Proceedings of the IEEE/CVF Winter Conference on
  Applications of Computer Vision (WACV)}, pages 98--107, 2022.

\bibitem[Hendrycks et~al.(2019)Hendrycks, Basart, Mazeika, Mostajabi,
  Steinhardt, and Song]{hendrycks_benchmark_anomaly_segmentation}
D.~Hendrycks, S.~Basart, M.~Mazeika, M.~Mostajabi, J.~Steinhardt, and D.~Song.
\newblock {A Benchmark for Anomaly Segmentation}.
\newblock \emph{arXiv preprint arXiv:1911.11132}, 2019.

\bibitem[Hong and Choe(2020)]{Hong_2020_DecentralizationLoss}
E.~Hong and Y.~Choe.
\newblock Latent feature decentralization loss for one-class anomaly detection.
\newblock \emph{IEEE Access}, 8:\penalty0 165658--165669, 2020.
\newblock \doi{10.1109/ACCESS.2020.3022646}.

\bibitem[Hu et~al.(2020)Hu, Yang, Xie, Rosa, Guo, Wang, Trigoni, and
  Markham]{hu2019randla}
Q.~Hu, B.~Yang, L.~Xie, S.~Rosa, Y.~Guo, Z.~Wang, N.~Trigoni, and A.~Markham.
\newblock {RandLA-Net: Efficient Semantic Segmentation of Large-Scale Point
  Clouds}.
\newblock \emph{Proceedings of the IEEE Conference on Computer Vision and
  Pattern Recognition}, 2020.

\bibitem[Hu et~al.(2021)Hu, Yang, Xie, Rosa, Guo, Wang, Trigoni, and
  Markham]{hu2021learning}
Q.~Hu, B.~Yang, L.~Xie, S.~Rosa, Y.~Guo, Z.~Wang, N.~Trigoni, and A.~Markham.
\newblock Learning semantic segmentation of large-scale point clouds with
  random sampling.
\newblock \emph{IEEE Transactions on Pattern Analysis and Machine
  Intelligence}, 2021.

\bibitem[Kehl et~al.(2016)Kehl, Milletari, Tombari, Ilic, and
  Navab]{wadim2016localrgbddescriptors}
W.~Kehl, F.~Milletari, F.~Tombari, S.~Ilic, and N.~Navab.
\newblock {Deep Learning of Local RGB-D Patches for 3D Object Detection and 6D
  Pose Estimation}.
\newblock In B.~Leibe, J.~Matas, N.~Sebe, and M.~Welling, editors,
  \emph{Computer Vision -- ECCV 2016}, pages 205--220, Cham, 2016. Springer
  International Publishing.
\newblock ISBN 978-3-319-46487-9.

\bibitem[Krizhevsky et~al.(2012)Krizhevsky, Sutskever, and
  Hinton]{krizhevsky2012imagenet}
A.~Krizhevsky, I.~Sutskever, and G.~E. Hinton.
\newblock {ImageNet Classification with Deep Convolutional Neural Networks}.
\newblock In \emph{Proceedings of the 25th International Conference on Neural
  Information Processing Systems - Volume 1}, pages 1097--1105, 2012.

\bibitem[Liu et~al.(2020)Liu, Li, Zheng, Karanam, Wu, Bhanu, Radke, and
  Camps]{Liu_2020_VisuallyExplaining}
W.~Liu, R.~Li, M.~Zheng, S.~Karanam, Z.~Wu, B.~Bhanu, R.~J. Radke, and
  O.~Camps.
\newblock Towards visually explaining variational autoencoders.
\newblock In \emph{Proceedings of the IEEE/CVF Conference on Computer Vision
  and Pattern Recognition (CVPR)}, 2020.

\bibitem[Menze et~al.(2015)Menze, Jakab, Bauer, Kalpathy-Cramer, Farahani,
  Kirby, et~al.]{brats2015}
B.~H. Menze, A.~Jakab, S.~Bauer, J.~Kalpathy-Cramer, K.~Farahani, J.~Kirby,
  et~al.
\newblock {The Multimodal Brain Tumor Image Segmentation Benchmark ({BRATS})}.
\newblock \emph{IEEE Transactions on Medical Imaging}, 34\penalty0
  (10):\penalty0 1993--2024, 2015.
\newblock \doi{10.1109/TMI.2014.2377694}.

\bibitem[Mishra et~al.(2020)Mishra, Piciarelli, and Foresti]{Mishra_2020_piade}
P.~Mishra, C.~Piciarelli, and G.~L. Foresti.
\newblock A neural network for image anomaly detection with deep pyramidal
  representations and dynamic routing.
\newblock \emph{International Journal of Neural Systems}, 30\penalty0
  (10):\penalty0 2050060, 2020.
\newblock \doi{10.1142/S0129065720500604}.

\bibitem[Moenning and Dodgson(2003)]{moenning2003_farthest_point_sampling}
C.~Moenning and N.~A. Dodgson.
\newblock {Fast Marching farthest point sampling}.
\newblock In \emph{Eurographics 2003 - Posters}. Eurographics Association,
  2003.
\newblock \doi{10.2312/egp.20031024}.

\bibitem[Pang et~al.(2021)Pang, Shen, Cao, and Hengel]{pang2021adreview}
G.~Pang, C.~Shen, L.~Cao, and A.~V.~D. Hengel.
\newblock Deep learning for anomaly detection: A review.
\newblock \emph{ACM Comput. Surv.}, 54\penalty0 (2), 2021.
\newblock ISSN 0360-0300.
\newblock \doi{10.1145/3439950}.

\bibitem[Paszke et~al.(2019)Paszke, Gross, Massa, Lerer, Bradbury, Chanan,
  Killeen, Lin, Gimelshein, Antiga, Desmaison, Kopf, Yang, DeVito, Raison,
  Tejani, Chilamkurthy, Steiner, Fang, Bai, and Chintala]{Paszke_2019_PyTorch}
A.~Paszke, S.~Gross, F.~Massa, A.~Lerer, J.~Bradbury, G.~Chanan, T.~Killeen,
  Z.~Lin, N.~Gimelshein, L.~Antiga, A.~Desmaison, A.~Kopf, E.~Yang, Z.~DeVito,
  M.~Raison, A.~Tejani, S.~Chilamkurthy, B.~Steiner, L.~Fang, J.~Bai, and
  S.~Chintala.
\newblock {PyTorch: An Imperative Style, High-Performance Deep Learning
  Library}.
\newblock In \emph{{Advances in Neural Information Processing Systems}},
  volume~32, 2019.

\bibitem[Potter et~al.(2020)Potter, Donohoe, Greene, Pribisova, and
  Donahue]{Potter_2020_pandanet}
K.~M. Potter, B.~Donohoe, B.~Greene, A.~Pribisova, and E.~Donahue.
\newblock {Automatic detection of defects in high reliability as-built parts
  using x-ray CT}.
\newblock In \emph{Applications of Machine Learning 2020}, volume 11511, pages
  120 -- 136. International Society for Optics and Photonics, SPIE, 2020.
\newblock \doi{10.1117/12.2570459}.

\bibitem[Reiss et~al.(2021)Reiss, Cohen, Bergman, and Hoshen]{Reiss_2021_PANDA}
T.~Reiss, N.~Cohen, L.~Bergman, and Y.~Hoshen.
\newblock Panda: Adapting pretrained features for anomaly detection and
  segmentation.
\newblock In \emph{2021 IEEE/CVF Conference on Computer Vision and Pattern
  Recognition (CVPR)}, pages 2805--2813, 2021.
\newblock \doi{10.1109/CVPR46437.2021.00283}.

\bibitem[Rippel et~al.(2021)Rippel, Chavan, Lei, and
  Merhof]{Rippel_2021_Gaussian}
O.~Rippel, A.~Chavan, C.~Lei, and D.~Merhof.
\newblock {Transfer Learning Gaussian Anomaly Detection by Fine-Tuning
  Representations}.
\newblock arXiv preprint arXiv:2108.04116, 2021.

\bibitem[Salehi et~al.(2021)Salehi, Sadjadi, Baselizadeh, Rohban, and
  Rabiee]{Salehi_2021_CVPR}
M.~Salehi, N.~Sadjadi, S.~Baselizadeh, M.~H. Rohban, and H.~R. Rabiee.
\newblock Multiresolution knowledge distillation for anomaly detection.
\newblock In \emph{Proceedings of the IEEE/CVF Conference on Computer Vision
  and Pattern Recognition (CVPR)}, pages 14902--14912, 2021.

\bibitem[Salti et~al.(2014)Salti, Tombari, and {Di
  Stefano}]{salti_2014_shot_descriptor}
S.~Salti, F.~Tombari, and L.~{Di Stefano}.
\newblock {SHOT: Unique signatures of histograms for surface and texture
  description}.
\newblock \emph{Computer Vision and Image Understanding}, 125:\penalty0
  251--264, 2014.
\newblock ISSN 1077-3142.
\newblock \doi{10.1016/j.cviu.2014.04.011}.

\bibitem[Schlegl et~al.(2019)Schlegl, Seeb{\"o}ck, Waldstein, Langs, and
  Schmidt-Erfurth]{Schlegl_2019_fAnoGan}
T.~Schlegl, P.~Seeb{\"o}ck, S.~M. Waldstein, G.~Langs, and U.~Schmidt-Erfurth.
\newblock {f-AnoGAN: Fast unsupervised anomaly detection with generative
  adversarial networks}.
\newblock \emph{Medical Image Analysis}, 54:\penalty0 30--44, 2019.
\newblock \doi{10.1016/j.media.2019.01.010}.

\bibitem[Simarro~Viana et~al.(2021)Simarro~Viana, de~la Rosa, Vande~Vyvere,
  Robben, Sima, and {CENTER-TBI Participants and
  Investigators}]{Simarro_Viana_2021}
J.~Simarro~Viana, E.~de~la Rosa, T.~Vande~Vyvere, D.~Robben, D.~M. Sima, and
  {CENTER-TBI Participants and Investigators}.
\newblock {Unsupervised 3D Brain Anomaly Detection}.
\newblock In \emph{Brainlesion: Glioma, Multiple Sclerosis, Stroke and
  Traumatic Brain Injuries}, pages 133--142. Springer International Publishing,
  2021.
\newblock \doi{10.1007/978-3-030-72084-1}.

\bibitem[Song and Yan(2013)]{song_steel_surface_defect_database}
K.~Song and Y.~Yan.
\newblock A noise robust method based on completed local binary patterns for
  hot-rolled steel strip surface defects.
\newblock \emph{Applied Surface Science}, 285:\penalty0 858--864, 2013.
\newblock \doi{10.1016/j.apsusc.2013.09.002}.

\bibitem[Tombari et~al.(2010)Tombari, Salti, and
  Di~Stefano]{tombari_2010_descriptors}
F.~Tombari, S.~Salti, and L.~Di~Stefano.
\newblock Unique signatures of histograms for local surface description.
\newblock In \emph{Proceedings of the 11th European Conference on Computer
  Vision: Part III}, page 356–369, Berlin, Heidelberg, 2010. Springer-Verlag.
\newblock ISBN 364215557X.

\bibitem[Venkataramanan et~al.(2020)Venkataramanan, Peng, Singh, and
  Mahalanobis]{Venkataramanan_2020_AttentionGuided}
S.~Venkataramanan, K.-C. Peng, R.~V. Singh, and A.~Mahalanobis.
\newblock Attention guided anomaly localization in images.
\newblock In A.~Vedaldi, H.~Bischof, T.~Brox, and J.-M. Frahm, editors,
  \emph{Computer Vision -- ECCV 2020}, pages 485--503. Springer International
  Publishing, 2020.
\newblock ISBN 978-3-030-58520-4.

\bibitem[Wang et~al.(2020)Wang, Zhang, Guo, and
  Han]{Wang_2020_LatentSpaceResampling}
L.~Wang, D.~Zhang, J.~Guo, and Y.~Han.
\newblock Image anomaly detection using normal data only by latent space
  resampling.
\newblock \emph{Applied Sciences}, 10\penalty0 (23), 2020.
\newblock ISSN 2076-3417.
\newblock \doi{10.3390/app10238660}.

\bibitem[Wu et~al.(2015)Wu, Song, Khosla, Yu, Zhang, Tang, and
  Xiao]{wu20153d_modelnet10}
Z.~Wu, S.~Song, A.~Khosla, F.~Yu, L.~Zhang, X.~Tang, and J.~Xiao.
\newblock {3D ShapeNets: A Deep Representation for Volumetric Shapes}.
\newblock In \emph{Proceedings of the IEEE conference on computer vision and
  pattern recognition}, pages 1912--1920, 2015.

\bibitem[Xie et~al.(2020)Xie, Gu, Guo, Qi, Guibas, and
  Litany]{PointContrast2020}
S.~Xie, J.~Gu, D.~Guo, C.~Qi, L.~Guibas, and O.~Litany.
\newblock {PointContrast: Unsupervised Pre-training for 3D Point Cloud
  Understanding}.
\newblock In \emph{ECCV}, 2020.

\bibitem[Zeng et~al.(2017)Zeng, Song, Nießner, Fisher, Xiao, and
  Funkhouser]{Zeng_2017_3DMatch}
A.~Zeng, S.~Song, M.~Nießner, M.~Fisher, J.~Xiao, and T.~Funkhouser.
\newblock {3DMatch: Learning Local Geometric Descriptors from RGB-D
  Reconstructions}.
\newblock In \emph{2017 IEEE Conference on Computer Vision and Pattern
  Recognition (CVPR)}, pages 199--208, 2017.
\newblock \doi{10.1109/CVPR.2017.29}.

\end{thebibliography}

\appendix

\clearpage

\section{Additional Information on Ablation Experiments}
\label{appendix:ablation}

For reference, we provide numerical values of the results of our ablation studies that correspond to the line- and barplots in our main manuscript.

\subsection{Lower Integration Limits}

In our paper, quantitative results are reported as the AU-PRO where the PRO values are integrated over the false positive rates (FPR). In the majority of our experiments, we limit the FPR by an upper integration limit of $0.3$. In order to enable a comparison at lower integration limits, we list the performance of our method at four different integration limits $\{0.01, 0.05, 0.10, 0.20\}$ in \Cref{table:supp_integration_limits}. The first four row show the performance of our model with a descriptor dimension of $d=64$. The bottom four rows show the corresponding performance of our model with feature dimension $d = 128$. These values are also depicted in the line plot shown in \Cref{fig:ablation_integration_limit}.

\begin{table}[h]
\centering
\resizebox{\columnwidth}{!}{%
\begin{tabular}{cc|cccccccccc|c}
\hline
 & \multicolumn{1}{c|}{\begin{tabular}[c]{@{}c@{}}integration\\ limit\end{tabular}} & bagel & \begin{tabular}[c]{@{}l@{}}cable\\ gland\end{tabular} & carrot & cookie & dowel & foam & peach & potato & rope & tire & mean \\ \hline \hline
\multirow{4}{*}{\rotatebox[origin=c]{90}{$d = 64$}} & 0.01 & 0.361 & 0.019 & 0.690 & 0.461 & 0.147 & 0.232 & 0.433 & 0.762 & 0.558 & 0.008 & 0.367 \\
 & 0.05 & 0.718 & 0.095 & 0.909 & 0.699 & 0.481 & 0.428 & 0.735 & 0.934 & 0.841 & 0.083 & 0.592 \\
 & 0.10 & 0.832 & 0.177 & 0.954 & 0.791 & 0.662 & 0.505 & 0.837 & 0.966 & 0.910 & 0.187 & 0.682 \\
 & 0.20 & 0.910 & 0.322 & 0.977 & 0.869 & 0.815 & 0.578 & 0.909 & 0.983 & 0.952 & 0.364 & 0.768 \\ \hline
\multirow{4}{*}{\rotatebox[origin=c]{90}{$d = 128$}} & 0.01 & 0.438 & 0.023 & 0.710 & 0.500 & 0.239 & 0.278 & 0.511 & 0.791 & 0.630 & 0.015 & 0.414 \\
 & 0.05 & 0.776 & 0.114 & 0.917 & 0.741 & 0.581 & 0.456 & 0.773 & 0.933 & 0.876 & 0.113 & 0.628 \\
 & 0.10 & 0.867 & 0.200 & 0.957 & 0.824 & 0.738 & 0.521 & 0.858 & 0.964 & 0.932 & 0.223 & 0.709 \\
 & 0.20 & 0.927 & 0.352 & 0.979 & 0.891 & 0.858 & 0.584 & 0.920 & 0.982 & 0.965 & 0.399 & 0.786 \\ \cline{1-13} 
\end{tabular}
}
\tabularspace
\caption{Performance of our method for various integration limits for each dataset category of the MVTec 3D-AD dataset.}
\label{table:supp_integration_limits}
\end{table}

\subsection{Varying Key Model Hyperparameters}

We analyzed the dependency of the anomaly detection performance of our method on the number of input points, the feature dimension of the geometric descriptors, and the number of nearest neighbors used for local feature aggregation. In \Cref{table:ablation_model_params}, we provide the numerical values for this ablation study. The values correspond to the line plots in \Cref{fig:ablation_study}.

\begin{table}[h]
\centering
\resizebox{\columnwidth}{!}{%
\begin{tabular}{l|cccc|ccccc|cccc}
 \hline
 Performance & \multicolumn{4}{c|}{Number of Points} & \multicolumn{5}{c|}{Feature Dimension} & \multicolumn{4}{c}{Nearest Neighbors} \\
 Metric & 16 & 32 & 64 & 128 & 16 & 32 & 64 & 128 & 256 & 8 & 16 & 32 & 64 \\ \hline \hline
\begin{tabular}[c]{@{}l@{}}Localization\\ (AU-PRO)\end{tabular} & 0.759 & 0.803 & 0.818 & 0.821 & 0.740 & 0.785 & 0.818 & 0.833 & 0.833 & 0.755 & 0.804 & 0.818 & 0.821 \\ \hline
\begin{tabular}[c]{@{}l@{}}Inference\\ Time (s)\end{tabular} & 0.014 & 0.026 & 0.049 & 0.080 & 0.017 & 0.028 & 0.049 & 0.093 & 0.189 & 0.020 & 0.030 & 0.049 & 0.080 \\ \hline
\begin{tabular}[c]{@{}l@{}}Training\\ Mem. (GB)\end{tabular} & 2.29 & 3.33 & 5.89 & 7.87 & 2.06 & 3.54 & 5.89 & 10.98 & 21.13 & 2.43 & 3.98 & 5.89 & 10.86 \\ \hline
\begin{tabular}[c]{@{}l@{}}Inference\\ Mem. (GB)\end{tabular} & 1.71 & 2.15 & 2.71 & 3.47 & 1.75 & 2.24 & 2.71 & 5.14 & 8.81 & 1.79 & 2.28 & 2.71 & 4.76 \\ \hline
\end{tabular}
}
\tabularspace
\caption{Ablation study on various hyperparameters of our proposed method.}
\label{table:ablation_model_params}
\end{table}

\subsection{Modifying the Training Strategy}

We further experimented with different training strategies applied to our proposed approach. We tested a randomly initialized teacher network, adding absolute point coordinates to the model, and incorporating rotation invariance to the anomaly detection training.  We then investigated different pretraining datasets and pretrained feature extractors. The numerical values for these experiments are listed in \Cref{table:ablation_training_strategy}. These values correspond to the bar plots in \Cref{fig:ablation_barplot}.

\begin{table}[h]
\centering
\begin{tabular}{lc|lc|lc}
\hline
\begin{tabular}[c]{@{}l@{}}Training\\ Strategy\end{tabular} & \multicolumn{1}{l|}{AU-PRO} & \begin{tabular}[c]{@{}l@{}}Pretraining\\ Dataset\end{tabular} & \multicolumn{1}{l|}{AU-PRO} & \begin{tabular}[c]{@{}l@{}}Feature\\ Extractor\end{tabular} & \multicolumn{1}{l}{AU-PRO} \\ \hline \hline
\begin{tabular}[c]{@{}l@{}}Random\\ Weights\end{tabular} & 0.278 & ITODD & 0.820 & FCGF & 0.719 \\ \hline
\begin{tabular}[c]{@{}l@{}}Absolute\\ Coordinates\end{tabular} & 0.505 & 3DMatch & 0.819 & PPF-FoldNet & 0.682 \\ \hline
\begin{tabular}[c]{@{}l@{}}Rotation\\ Invariance\end{tabular} & 0.650 & KITTI & 0.735 & \multicolumn{1}{c}{-} & -  \\ \hline
\end{tabular}
\tabularspace
\caption{Performance of our method when making various changes to the training strategy. The AU-PRO is reported curve up to an integration limit of 0.3.}
\label{table:ablation_training_strategy}
\end{table}

\clearpage
\section{Additional Qualitative Results}
\label{appendix:qualitatives}

\begin{figure}[h]
    \centering
    \includegraphics[width=0.98\textwidth]{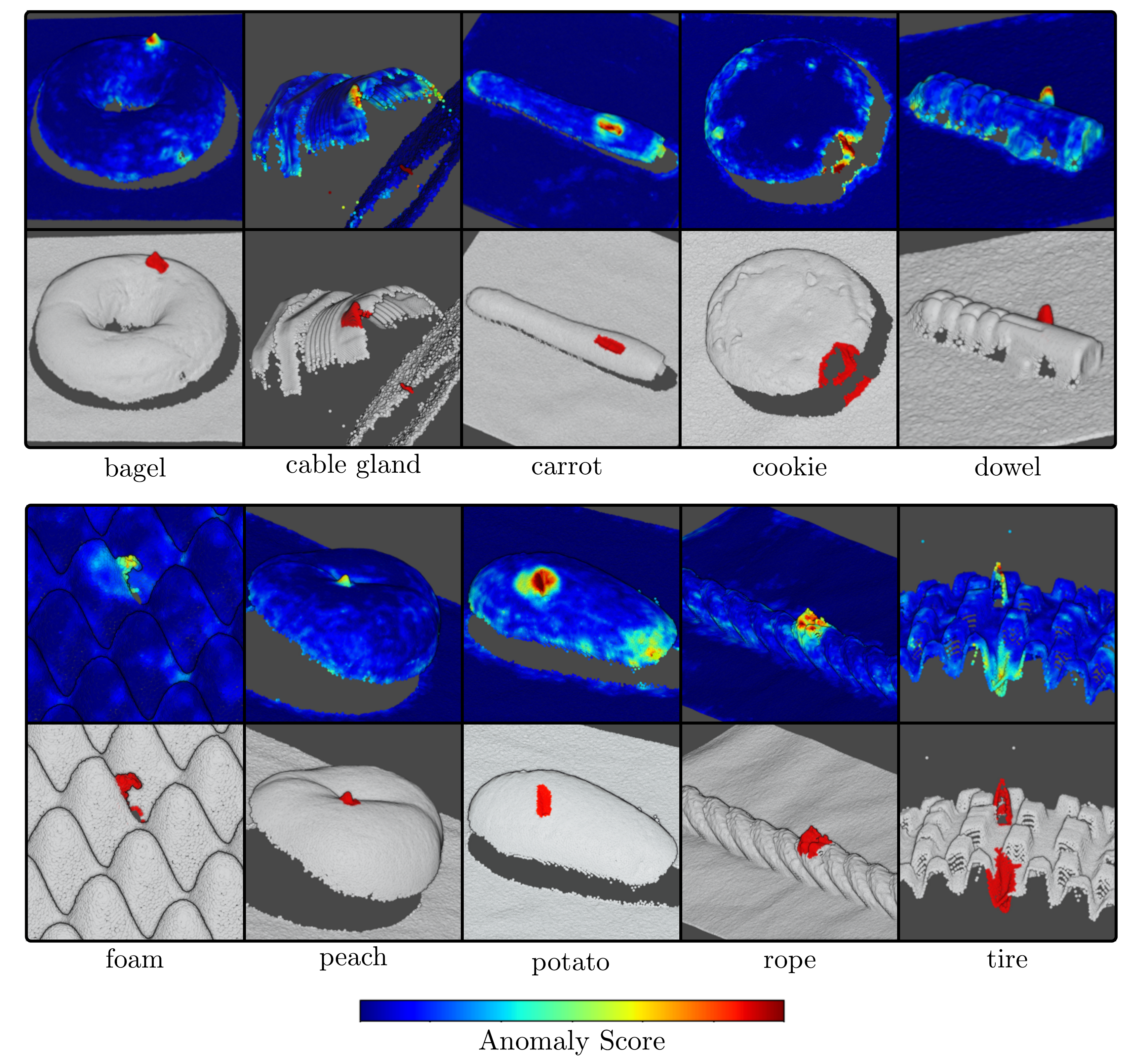}
    \caption{Additional qualitative results of our method on the MVTec 3D-AD dataset. Top row: Anomaly scores for each 3D point predicted by our algorithm. Bottom row: Ground truth annotations of anomalous points in red.}
    \label{fig:qualitatives_appendix}
\end{figure}

\Cref{fig:qualitatives_appendix} shows additional qualitative results of our method for each dataset category of the MVTec 3D-AD dataset for which our method reliably localizes anomalies.

\end{document}